\documentclass[journal]{IEEEtran}
\usepackage{rotating}
\usepackage{booktabs}
\usepackage{hyperref}
\usepackage[hyphenbreaks]{breakurl}

\usepackage{amsmath}

\usepackage{array}

\begin{document}

\title{Toward more accurate and generalizable brain deformation estimators for traumatic brain injury detection with unsupervised domain adaptation}

\author{{Xianghao Zhan, Jiawei Sun, Yuzhe Liu, Nicholas J. Cecchi, Enora Le Flao, Olivier Gevaert, Michael M. Zeineh, David B. Camarillo}

\thanks{Xianghao Zhan, Jiawei Sun, Yuzhe Liu, Nicholas J. Cecchi, Enora Le Flao, David B. Camarillo are with Department of Bioengineering, Stanford University, CA, 94305, USA; Yuzhe Liu is with School of Biological Science and Medical Engineering, BeiHang University, Beijing, 10019, China; Olivier Gevaert is with Department of Biomedical Data Science, Stanford University, CA, 94305, USA; Michael M. Zeineh is with Department of Radiology, Stanford University, CA, 94305, USA. This work was supported by the Pac-12 Conference’s Student-Athlete Health and Well-Being Initiative, the National Institutes of Health (R24NS098518), Taube Stanford Children’s Concussion Initiative and Stanford Department of Bioengineering. (Corresponding Authors: Yuzhe Liu e-mail: yuzheliu@buaa.edu.cn)}}%

\markboth{}%
{Shell \MakeLowercase{\textit{et al.}}: Bare Demo of IEEEtran.cls for IEEE Journals}

\maketitle

\begin{abstract}
Machine learning head models (MLHMs) are developed to estimate brain deformation for early detection of traumatic brain injury (TBI). However, the overfitting to simulated impacts and the lack of generalizability caused by distributional shift of different head impact datasets hinders the broad clinical applications of current MLHMs. We propose brain deformation estimators that integrates unsupervised domain adaptation with a deep neural network to predict whole-brain maximum principal strain (MPS) and MPS rate (MPSR). With 12,780 simulated head impacts, we performed unsupervised domain adaptation on on-field head impacts from 302 college football (CF) impacts and 457 mixed martial arts (MMA) impacts using domain regularized component analysis (DRCA) and cycle-GAN-based methods. The new model improved the MPS/MPSR estimation accuracy, with the DRCA method significantly outperforming other domain adaptation methods in prediction accuracy ($p<0.001$): MPS RMSE: 0.027 (CF) and 0.037 (MMA); MPSR RMSE: 7.159 (CF) and 13.022 (MMA). On another two hold-out test sets with 195 college football impacts and 260 boxing impacts, the DRCA model significantly outperformed the baseline model without domain adaptation in MPS and MPSR estimation accuracy ($p<0.001$). The DRCA domain adaptation reduces the MPS/MPSR estimation error to be well below TBI thresholds, enabling accurate brain deformation estimation to detect TBI in future clinical applications.
\end{abstract}


\IEEEpeerreviewmaketitle

\section{Introduction}

%1. TBI & mTBI & monitoring
%Traumatic brain injury (TBI) has become a global health issue affecting more than 55 million people worldwide \cite{james2019global}. TBI has also been one of the leading causes of disability and death worldwide, particularly for children and young adults \cite{maas2008moderate}. TBI can be caused by head impacts from various sources including traffic accidents, battlefield blasts, various contact sports, and domestic abuse \cite{hernandez2015six,o2020dynamic,cecchi2019head}. The neuropathological classification groups TBI into two types \cite{o2011animal}: 1) primary injury, which is the brain damage caused directly and instantly by the injury, such as contusions, intracranial hemorrhage, and traumatic axonal injury; 2) secondary injury, which is the brain damage caused by prolonged pathological processes initiated by impact, such as neuroinflammation and blood-brain barrier disruption. Mainly due to the existence of secondary injury, repetitive incidence of mild TBI (mTBI) can contribute to brain damage accumulation. Repetitive mTBI can induce long-term cognitive deficits and even neurodegeneration \cite{doherty2016blood}, and the early detection of TBI has proven essential to prevention and recovery of brain damage. This underscores the need for proper monitoring, as quick diagnosis and early intervention of brain damage \cite{beckwith2013timing} is a crucial step to prevent repetitive mTBI \cite{ponsford2001impact,guiza2017early}.

%2. Brain strain, strain rate, FEM, deep learning models
In traumatic brain injury (TBI), the brain is damaged by tissue deformation under the inertial loadings when the skull sustains high acceleration or deceleration. Therefore, metrics of brain deformation are effective predictors of TBI. Brain strain, particularly the maximum principal strain (MPS), has been shown to correlate with TBI pathologies \cite{o2020dynamic,hajiaghamemar2020head,hajiaghamemar2021multi}, and the maximum principal strain rate (MPSR) has been found to correlate with traumatic axonal injury \cite{hajiaghamemar2020head,hajiaghamemar2020embedded} and effectively in predicting TBI pathologies across species \cite{wu2021evaluation,wu2022integrating}. To compute tissue-level strain and strain rate, finite element (FE) head model, which takes head kinematics as inputs, is regarded as the state-of-the-art approach\cite{mao2013development,kleiven2007predictors}, but its applicability to a broader user community is limited by its long computation times (approximately 7-8 hours per impact \cite{zhan2021rapid}) and complicated finite element modeling software. Recently developed machine learning head models (MLHMs) trained on large quantities of FE head model simulations, in contrast, can substantially reduce the computational cost (to be within 1ms per impact reported with a personal computer \cite{zhan2021rapid}), enabling TBI detection that uses brain strain and strain rate as biomechanics-based quantitative metrics for TBI risks\cite{wu2019convolutional,zhan2021rapid,ghazi2021instantaneous,wu2022real,zhan2022find}.

%3. Accuracy decreasing problem when types are different
Despite their superiority in computational time, MLHMs are often biased towards data similar to the distribution of their training data due to overfitting, which leads to a decreasing accuracy when the test data is distinctive from the training data. \cite{zhan2021rapid,zhan2021predictive,zhan2021relationship}. To solve this issue, MLHMs were developed for specific head impact types, and classification and cluster models were developed to identify the head impact type to select the adequate MLHM to predict MPS and MPSR \cite{zhan2021classification,zhan2021clustering}. A limitation of this solution is the availability of head impact data for specific types. In a previous study \cite{zhan2021rapid}, promising accuracy was found using a simulation dataset of 2,130 impacts, while only several hundred or even less than a hundred head impacts are available in some types of head impacts, e.g. NASCAR, wrestling \cite{zhan2021predictive,zhan2021relationship}. Directly combining datasets as the training dataset will lead to a significant reduction in accuracy\cite{zhan2021rapid}, so Principal Component Analysis was applied to simplify the structure of the deep learning network and reduce the size of the training dataset needed to develop MLHM just based on data from specific head impact type. \cite{zhan2022find}.

%Domain Adaptation
Identifying head impact type and predicting with specific MLHM solved the issue of decreasing accuracy, but it requires developing classifiers and MLHMs for every head impact type \cite{zhan2021classification,zhan2022find}, which is unlikely doable considering the number of head impact types in the realistic. Therefore, we propose domain adaptation as a new approach to solving the issue. The domain adaptation enables an automatic model configuration based on the data collected from a new head impact type, and then configured model can predict the MPS and MPSR for this head impact type. In domain adaptation, the source domain is a large dataset consisting of 12,780 head impacts covering all potential head impact directions, locations, and speeds using FE simulations, and the data collected from the new head impact type is considered the target domain. The drift between the source and target domains leads to the decreasing accuracy of MLHM, so the configuration before predicting adapts the target domain to the source domain, which compensates for the drift and improves the prediction accuracy.

The configuration of domain adaptation enables the application of MLHMs to a new type of head impacts without sacrificing accuracy nor developing new MLHM with supervised model finetuning by running a large amount of FE simulations. This work will accelerate the application of MLHM in the various head impact scenarios and ultimately contribute to the improvement of TBI diagnosis and protection.

	\begin{table}
		\centering
		\caption{The acronyms and abbreviations used in this paper and their respective meanings.}
		\label{tab:1}       % Give a unique label
		\begin{tabular}{cc}
			\hline\noalign{\smallskip}
			Acronym/Abbreviation & Meaning  \\
			\noalign{\smallskip}\hline\noalign{\smallskip}
			MLHM & machine learning head model \\
			TBI & traumatic brain injury \\
	        MPS & maximum principal strain \\ 
	        MPSR & maximum principal strain rate\\ 
	        MMA & mixed martial arts \\
	        CF & college football \\
	        DRCA & domain regularized component analysis \\
	        GAN & generative adversarial network \\
	        mTBI & mild traumatic brain injury \\
            KMM & kernel mean matching \\
            HM & head model simulated dataset \\
            Ang. Vel. Mag. & angular velocity magnitude \\
            Ang. Acc. Mag. & angular acceleration magnitude \\
            PCA & principal component analysis \\
            t-SNE & t-distributed stochastic neighbor embedding \\
            DNN & deeo neural network \\
	        ReLu & rectified linear unit \\
	        Adam & adaptive moment estimation \\
	        MAE &  mean absolute error \\
	        RMSE & root mean squared error \\
			\hline\noalign{\smallskip}
		\end{tabular}
	\end{table}

%%%%%%%%%%%%%%%%%%%%%%%
\section{Methods}

\subsection{Datasets}
To provide enough data to train the model, we generated a large dataset consisting of 2,130 head impacts using FE models of a pneumatic impactor and hybrid III headform \cite{giudice2019development}, and then augmented the dataset by switching axes of kinematics \cite{wu2022integrating,wu2022real,ghazi2021instantaneous}. The augmented dataset had 12,780 head impacts, which was enough to train the model compared with previous studies \cite{zhan2021rapid,wu2022real,ghazi2021instantaneous}, and was then considered the source domain (denoted as HM). 

Two previously published datasets were used to tune the hyperparameters in the domain adaption model: college football head impacts(denoted as CF1) \cite{hernandez2015six}, 457 MMA head impacts (denoted as MMA) \cite{o2020dynamic,tiernan2020concussion}. The data were collected with Stanford Instrumented Mouthguard \cite{liu2020validation,camarillo2013instrumented}. The datasets CF1 and MMA are used as the validation datasets to develop the unsupervised domain adaptation models and to select the domain adaptation model adopted in configuration.

After selecting the best-performance domain adaptation model, another two datasets were used to perform additional hold-out test: 1) 195 recently collected college football head impacts by Stanford instrumented mouthguards (denoted as CF2, this dataset has not been published); 2) 260 boxing head impacts using the Prevent Biometrics Hybrid mouthguards \cite{le2022head} (dataset Boxing).

\subsection{Feature engineering}
The head kinematics in each dataset were represented by a $N \times D$ feature matrices with $N$ denoting the sample size and $D$ denoting the number of kinematic features. From each sample, we extracted 510 temporal and spectral features according to four physical quantities describing the head translation and rotation: linear acceleration at the brain's center of gravity $a(t)$, angular velocity $\omega(t)$, angular acceleration $\alpha(t)$, and angular jerk $j(t)$. Four channels are associated with each quantity: the three anatomical axes (x: posterior-to-anterior, y: left-to-right, z: superior-to-inferior) and the magnitude. The reason to include the temporal features was because of their high predictability of brain strain \cite{zhan2021rapid,zhan2021predictive}. The temporal features were able to explain a high variance of the 95th percentile MPS in linear models \cite{zhan2021predictive}. The reason to include the spectral features was that based on the spectral features several basic classifiers were able to reach high accuracy in classifying different types of head impacts, which indicated that the spectral features carry much information associated with the types of head impacts (the domain information in this study) \cite{zhan2021classification}. Therefore, to develop models that are able to generalize across different domains, we included the spectral features in our feature set. The details of the feature extraction can be found in the publications of previously developed MLHMs, as it was demonstrated that the extracted features are sufficient for developing an accurate machine learning head model for prediction of the MPS and MPSR \cite{zhan2021rapidly,zhan2021rapid}.

The reference MPS and MPSR are computed with a validated finite element head model: the KTH finite element model (Stockholm, Sweden) \cite{kleiven2007predictors}, which modeled the human brain with 4,124 brain elements, covering the brain, skull, scalp, meninges, falx, tentorium, subarachnoid cerebrospinal fluid (CSF), ventricles, and 11 pairs of the largest bridging veins. The MPS and MPSR of brain elements are used as labels to train our models and quantify prediction accuracy. 

\begin{figure*}[!ht]
    \centering
    \includegraphics[width=0.99\linewidth]{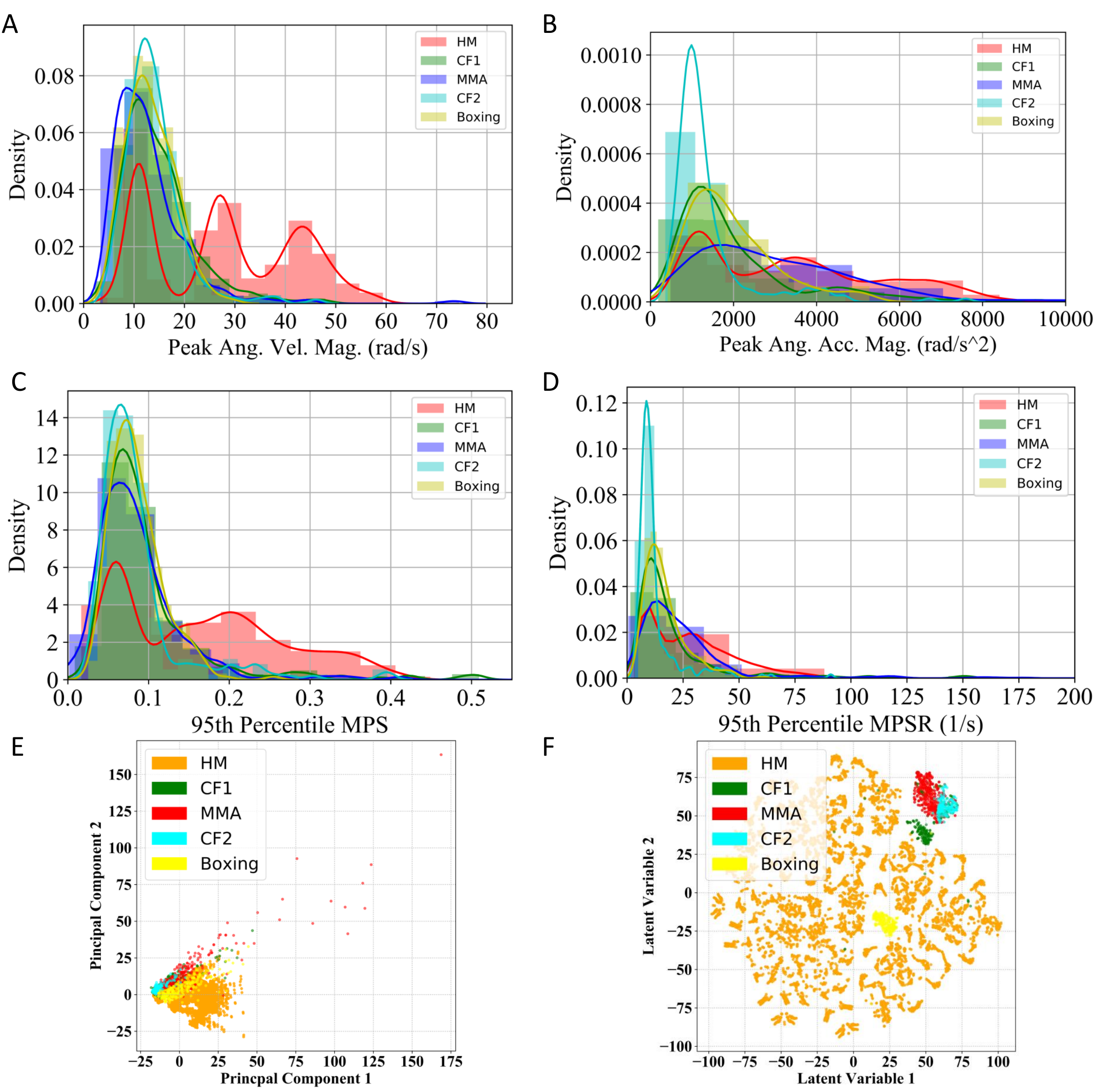}
    \caption{Distribution of MPS and MPSR of HM, CF1, MMA, CF2 and Boxing datasets. The distribution of the peak angular velocity magnitude (peak Ang. Vel. Mag., A), the peak angular acceleration magnitude (peak Ang. Acc. Mag. B), the 95th percentile MPS (C) and the distribution of the 95th percentile MPSR (D). The principal component analysis (PCA) visualization (E) and t-distributed stochastic neighbor embedding (t-SNE) visualization of all types of head impacts (F). Note: 95th percentile is frequently used in TBI biomechanics research to avoid numerical instability in the maximum values.}
    \label{fig:1}
\end{figure*}

Given the diverse sources of our impact data, the difference among datasets can be visualized by the distribution of the peak rotational kinematics (Fig. \ref{fig:1}A, B), the distribution of the 95th percentile MPS and MPSR (Fig. \ref{fig:1}C, D) and the visualization of the kinematics features with dimensionality reduction approaches (Fig. \ref{fig:1}E, F), underlining the challenges of obtaining a generalizable model.

\subsection{Structure and Training of Machine Learning Head Model}
\begin{figure}[!ht]
    \centering
    \includegraphics[width=0.99\linewidth]{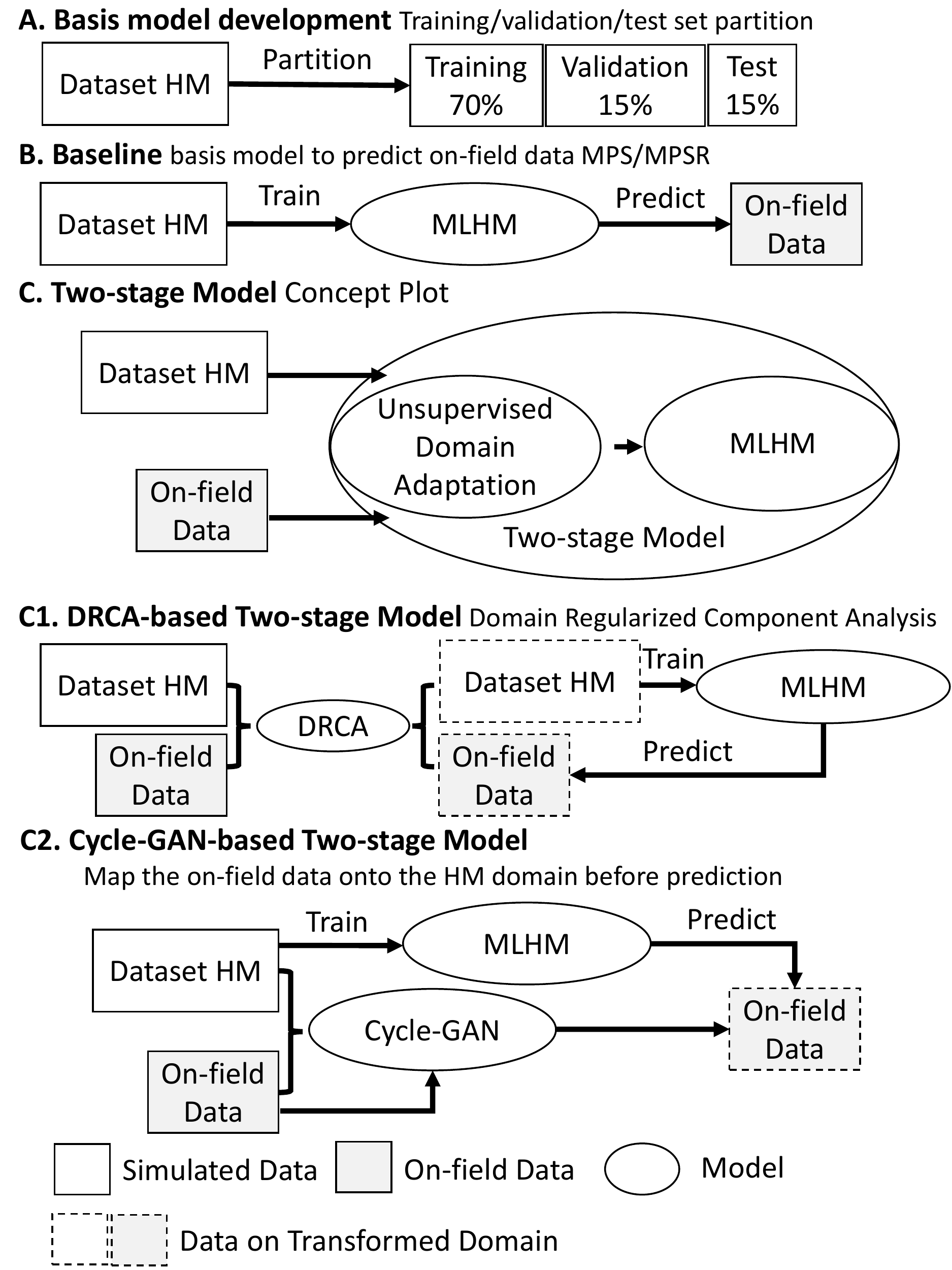}
    \caption{The pipeline of the model development. The pipeline consists of a deformation prediction model and domain adaptation performed by DRCA or cycle-GAN. Specifically, it shows the development of basis model on dataset HM (A), the baseline method (B), the DRCA method (C), and the Cycle-GAN method (D).}
    \label{fig:2}
\end{figure}

%\subsection{Brain Deformation Prediction Modeling}
DNN models are trained as the deformation estimation models that map kinematic features to MPS and MPSR, respectively, using a previously validated protocol \cite{zhan2021rapid}. The model is trained solely on the dataset HM where the dataset size suffices to yield a reasonably accurate deep neural network (Fig. \ref{fig:2}A). The DNN consists of five layers in addition to the input layer (510 neurons) and the output layer (4124 neurons): 1) hidden layer 1: 500 neurons (activation: the rectified linear unit (ReLU)); 2) dropout layer 1 (dropout rate 0.5); 3) hidden layer 2: 300 neurons (activation: ReLU); 4) dropout layer 2 (dropout rate 0.5); 5) hidden layer 3: 100 neurons (activation: ReLU). The design of this structure followed the guidance of firstly condensing kinematics information from the 510 features and then predicting the output variable. The loss function is the mean squared error coupled to an $L_2$ regularization term to avoid overfitting. To train the model, the entire dataset HM (12,780 impacts) was randomly split into 70\% training data for model training, 15\% validation data for hyperparameter tuning and 15\% test data for model performance evaluation.

\subsection{Domain Adaptation and MLHM Configuration}

\subsubsection{Baseline}
As a baseline, the MLHM trained on the HM dataset is directly used to predict the target datasets: CF1, MMA, CF2 and Boxing datasets (Fig. \ref{fig:2}B).

\subsubsection{Domain regularized component analysis (DRCA) Domain Adaptation}
DRCA \cite{anti-drift} converts the kinematics features to a feature subspace where the scatter between the source and target domains is minimized while the scatter in each domain is retained. Then, the MLHM is configured by training the model on the HM dataset converted to the feature subspace. Because the hyperparameters in both DRCA and model training are determined on the validation datasets, the configuration can be performed automatically when the MLHM is applied to a new dataset. The first two variables in the subspace of DRCA for all datasets in this paper were plotted in Fig. \ref{fig:2}C.

%To perform domain adaptation, DNN models with the same architecture as in the baseline are trained by the simulated data converted by DRCA \cite{anti-drift} to a new feature subspace corresponding to the on-field impact data. After we projected the source domain (simulated data) and the target domain (on-field data) onto the regularized domain, according to Fig. \ref{fig:2}C, we trained the DNN models with the simulated data on the new domain and predict the MPS and MPSR for the on-field data.

DRCA \cite{anti-drift} is briefly reviewed as follows: DRCA looks for a projection hyper-plane on which the difference between the projected source domain (the dataset HM) and target domain (the dataset CF1/MMA) is minimized. If we denote the samples on the source domain as $x_i^S \in {\rm I\!R}^{D}$, $i=1,2,...,N^S$ and the ones on the target domain as $x_i^T \in {\rm I\!R}^{D}$, $i=1,2,...,N^T$, where $D$ is the original feature dimension, the statistics of the data distribution can be calculated as follows.

a) The mean of the source-domain data: $\mu^S = \frac{ \Sigma_{i=1}^{N^S}x_i^S}{N^S} \in \rm I\!R^{D}$, of the target-domain data: $\mu^T = \frac{\Sigma_{i=1}^{N^T}x_i^T}{N^T}  \in \rm I\!R^{D}$, and of these data combined: $\mu = \frac{N^S \times \mu^S + N^T \times \mu^T}{N^S+N^T}$;

b) The within-domain scatter of the source-domain data and of the target-domain data: $S_w^S = \Sigma_{i=1}^{N^S}(x_i^S-\mu^S)(x_i^S-\mu^S)^T \in \rm I\!R^{D \times D}$ and $S_w^T= \Sigma_{i=1}^{N^T}(x_i^T-\mu^T)(x_i^T-\mu^T)^T \in \rm I\!R^{D \times D}$;

c) The between-domain scatter of the source-domain data and of the target-domain data: $S_b = N^S \times (\mu^S - \mu)(\mu^S - \mu)^T + N^T \times (\mu^T - \mu)(\mu^T - \mu)^T  \in \rm I\!R^{D \times D}$

The goal of the DRCA is to find a projection matrix $P \in {\rm I\!R}^{D \times d}~(d<D)$ to reduce the between-domain scatter while maintaining the within-domain scatter on the projected hyper-plane. Suppose the sample $x_i$ is projected onto $\Tilde{x_i}$:  $\Tilde{x_i} = P^T x_i \in \rm I\!R^{d} $. Then, the same three types of summary statistics can be calculated on the projected data. For example, the within domain scatter on the projection hyper-plane is represented as: $\Tilde{S_w^S} = P^TS_w^SP \in {\rm I\!R}^{d \times d}$ and $\Tilde{S_w^T} = P^TS_w^TP \in {\rm I\!R}^{d \times d}$. The between domain scatter on the projection hyper-plane is represented as: $\Tilde{S_b} = P^TS_bP \in {\rm I\!R}^{d \times d}$. Therefore, to minimize the domain difference while maintaining the data spread from each domain of data on the projection hyper-plane, we have the problem:
\begin{equation}
    \mathrm{max}_P \frac{tr(P^TS_w^SP + \alpha P^TS_w^TP)}{tr(P^TS_bP)} \mathrm{,}
\end{equation}
where $\alpha$ is a hyperparameter. The convex optimization problem in the fractional format can be reformulated as a constrained optimization problem. With a Lagrangian multiplier, the Lagrangian can be represented as:
\begin{equation}
    L(P, \theta) = tr(P^T S_w^S P + \alpha P^T S_w^T P) - \theta (tr(P^T S_b P) - \lambda)
\end{equation}

By taking the derivative of the Lagrangian with respect to $P$ and setting it to zero, the problem can be solved as an eigenvalue decomposition problem: $S_b^{-1}(S_w^S + \alpha S_w^T) P = \theta P$, where $P$ is the eigenvector matrix. By ranking the eigenvectors based on the eigenvalues, we extract the $d$ eigenvectors associated with the $d$ largest eigenvalues. Projected onto these eigenvectors, the data matrix is transformed from $N^S \times D$ and $N^T \times D$ to $N^S \times d$ and $N^T \times d$.

\subsubsection{Cycle-GAN-based Domain Adaptation}
Cycle-GAN was used to map head impact data from the target domain to the source domain (Fig. \ref{fig:2}D). Because the features were mapped to the same space, instead of the features subspace, the MLHM in the cycle-GAN and the baseline are the same. In the configuration, the cycle-GAN model is trained to transfer the data from the target domain to the source domain, so MLHM does not need to be trained in the configuration as DRCA and can be preset. 

Cycle-GAN is briefly reviewed as follows: Cycle-GAN model contains two generators ($G_t$, $G_s$) and two discriminators ($D_t$, $D_s$). The loss function is defined as a combination of the cycle-consistency losses and the classification losses. The noise $z$ was added by randomly dropping out certain dimension of features with probability $p$. Details of the structure optimization can be found in the supplementary materials.
\begin{equation}
\begin{aligned}
    L = \lambda_s\|G_t(G_s(x_s,z),z')-x_s\| + \lambda_t\|G_s(G_t(x_t,z'),z)-x_t\| \\
    - D_s(G_t(x_t,z')) - D_t(G_s(x_s,z))
\end{aligned}
\end{equation}
where $z$ and $z'$ are random vectors from $\mathcal{N}(0,I)$, and $\lambda_s$ and $\lambda_t$ are weight factors. The first two terms are the cycle-consistency losses and the last two terms are the classification losses. The optimization is defined as a minimax problem $\min_{G_s,G_t}\max_{D_s,D_t} L$.

\subsubsection{Shift-GAN-based Domain Adaptation}

To further reduce the potential domain drift between source and target domains after cycle-GAN-based adaptation, we applied shift-GAN, in which the Kernel Means Matching (KMM) is performed on the target dataset transferred by cycle-GAN to further minimize the discrepancy of the data in kernel space \cite{GAN}. Similar to the configuration of cycle-GAN, only the target dataset is transferred, so the MLHM can be preset.

\subsubsection{DRCA-GAN-based Domain Adaptation}

Furthermore, DRCA was performed based on the dataset transferred by cycle-GAN to see if it improve the performance. Because DRCA's outputs are in the subspace, both the cycle-GAN model and the MLHM need to be trained in the configuration.

%or the DRCA on the top of the cycle-GAN, which are compared against the cycle-GAN alone, the DRCA method, and the baseline without any domain adaptation. It is worth noting that the combination of cycle-GAN and KMM forms the protocol of shift-GAN previously proposed \cite{GAN}.

\subsection{Model performance validation and hold-out test}
On the two validation datasets (dataset CF1 and dataset MMA), we validated the feasibility of improving the accuracy of MPS and MPSR prediction and selected the domain adaptation model. The domain adaptation model and the subsequent MLHM were trained, validated, and tested on these two datasets. Then, the best domain adaptation model was selected by comparing the mean absolute error (MAE) of MPS and MPSR. The hyperparameters were tuned in this process, including the dimensionality of the projection hyperplane $d$ and relative weight on the target domain data $\alpha$ for DRCA; the number of layers, the number of neurons, the weight in the loss function $\lambda_s$ and $\lambda_t$, the number of epochs for the cycle-GAN; the number of layers, the number of neurons, the learning rate and the number of epochs for the MLHM.

Upon optimizing the hyperparameters and selecting the domain adaptation approach. We performed the additional model performance test on the two hold-out test datasets (CF2 and Boxing) with the hyperparameters fixed after the model validation process on the hold-out test datasets. The model performance in MAE and root mean squared error (RMSE) was reported in the results section.

\subsection{Statistical Test}
To evaluate the changes in model accuracy when compared with the baseline method, on the hold-out test datasets, we computed the MAE between the reference results given by the KTH finite element model and the predictions given by the tuned models on all the test impacts. To test the statistical significance, the paired t-test was used.

%%%%%%%%%%%%%%%%%%%%%%%%%%
\section{Results}
To quantify the overall accuracy of the prediction, we computed the MAE and RMSE of MPS and MPSR over all brain elements and then calculated the mean over all test impacts. According to the results shown in Table 2, in terms of the MAE, for the MPS prediction, the DRCA method reduces the MAE from 0.036 to 0.017 for dataset CF1, from 0.103 to 0.020 for dataset MMA; for MPSR prediction, the DRCA method reduces the MAE from 6.005 to 4.094 for dataset CF1 and from 1577.2 to 6.086 for dataset MMA. When compared with the previously reported model performance on the simulation dataset, the MAE in the MPS prediction was in the same level as the MAE reported in our previous studies when the model predicts the MPS on the test impacts originated from dataset HM without any domain drift (MAE: 0.015), on which the models were trained with the same kinematics features as in the current study \cite{zhan2021rapidly}. In terms of the RMSE, for the MPS prediction, the DRCA method reduces the mean RMSE from 0.063 to 0.027 (MPS, dataset CF1), from 1.582 to 0.037 (MPS, dataset MMA), from 12.449 to 7.159 (MPSR, datset CF1), and from over $10^{5}$ to 13.022 (MPSR, dataset MMA). With DRCA, the MPS prediction error was approximately 1/10 the presumed human concussion (mild TBI) thresholds (0.3-0.4 \cite{patton2015biomechanical,ho2007dynamic,kleiven2007predictors}) and the MPSR prediction error was also much smaller than the threshold for accurate axonal injury prediction in large animal models (120$s^{-1}$) \cite{hajiaghamemar2020embedded}, which suggested the capacity of MLHM with DRCA to detect the injury cases in on-field dataset.

\begin{table*}
		\centering
		\caption{The model performance metrics of the MPS and MPSR estimation with the baseline method and different domain adaptation approaches.}
		\label{tab:2}       % Give a unique label
		\begin{tabular}{ccccccc}
			\hline\noalign{\smallskip}
			Target & Dataset & Method & MAE & RMSE & Relative & Relative \\
                       & & & & & MAE Change & RMSE Change \\
			\noalign{\smallskip}\hline\noalign{\smallskip}
			MPS & CF1 & Baseline & 0.036 & 0.063 & 0 & 0\\
			MPS & CF1 & DRCA & 0.017 & 0.027 & -52.8\% & -57.1\%\\
			MPS & CF1 & Cycle-GAN & 0.037 & 0.052 & +2.8\% & -17.5\% \\
			MPS & CF1 & Shift-GAN & 0.034 & 0.049 & -5.6\% & -22.2\%\\
			MPS & CF1 & Cycle-GAN+DRCA & 0.032 & 0.050 & -11.1\% & -20.6\% \\
			MPSR & CF1 & Baseline & 6.005 & 12.449 & 0 & 0\\
			MPSR & CF1 & DRCA & 4.094 & 7.159 & -31.8\% & -42.5\%\\
			MPSR & CF1 & Cycle-GAN & 8.406 & 12.702 & +40.0\% & +2.0\%\\
			MPSR & CF1 & Shift-GAN & 8.019 & 13.301 & +33.5 \% & +6.8\%\\
			MPSR & CF1 & Cycle-GAN+DRCA & 7.764 & 13.854 & +29.3\% & +11.3\% \\
			MPS & MMA & Baseline & 0.103 & 1.582 & 0 & 0\\
			MPS & MMA & DRCA & 0.020 & 0.037 & -80.6\% & -97.7\%\\
			MPS & MMA & Cycle-GAN & 0.038 & 0.054 & -63.1\% & -96.6\% \\
			MPS & MMA & Shift-GAN & 0.041 & 0.054 & -60.2\% & -96.6\%\\
			MPS & MMA & Cycle-GAN+DRCA & 0.069 & 0.089 & -33.0\% & -94.4\%\\
			MPSR & MMA & Baseline & 1577.2 & 130157.7 & 0 & 0\\
			MPSR & MMA & DRCA & 6.610 & 12.206 & -99.6\% & -100.0\% \\
			MPSR & MMA & Cycle-GAN & 10.328 & 15.916 & -99.3\% & -100.0\% \\
			MPSR & MMA & Shift-GAN & 8.872 & 17.455 & -99.4\% & -100.0\% \\
			MPSR & MMA & Cycle-GAN+DRCA & 13.489 & 23.125 & -99.1\% & -100.0\%\\
			
			\\
			\hline\noalign{\smallskip}
		\end{tabular}
	\end{table*}

\begin{figure*}
    \centering
    \includegraphics[width=0.99\linewidth]{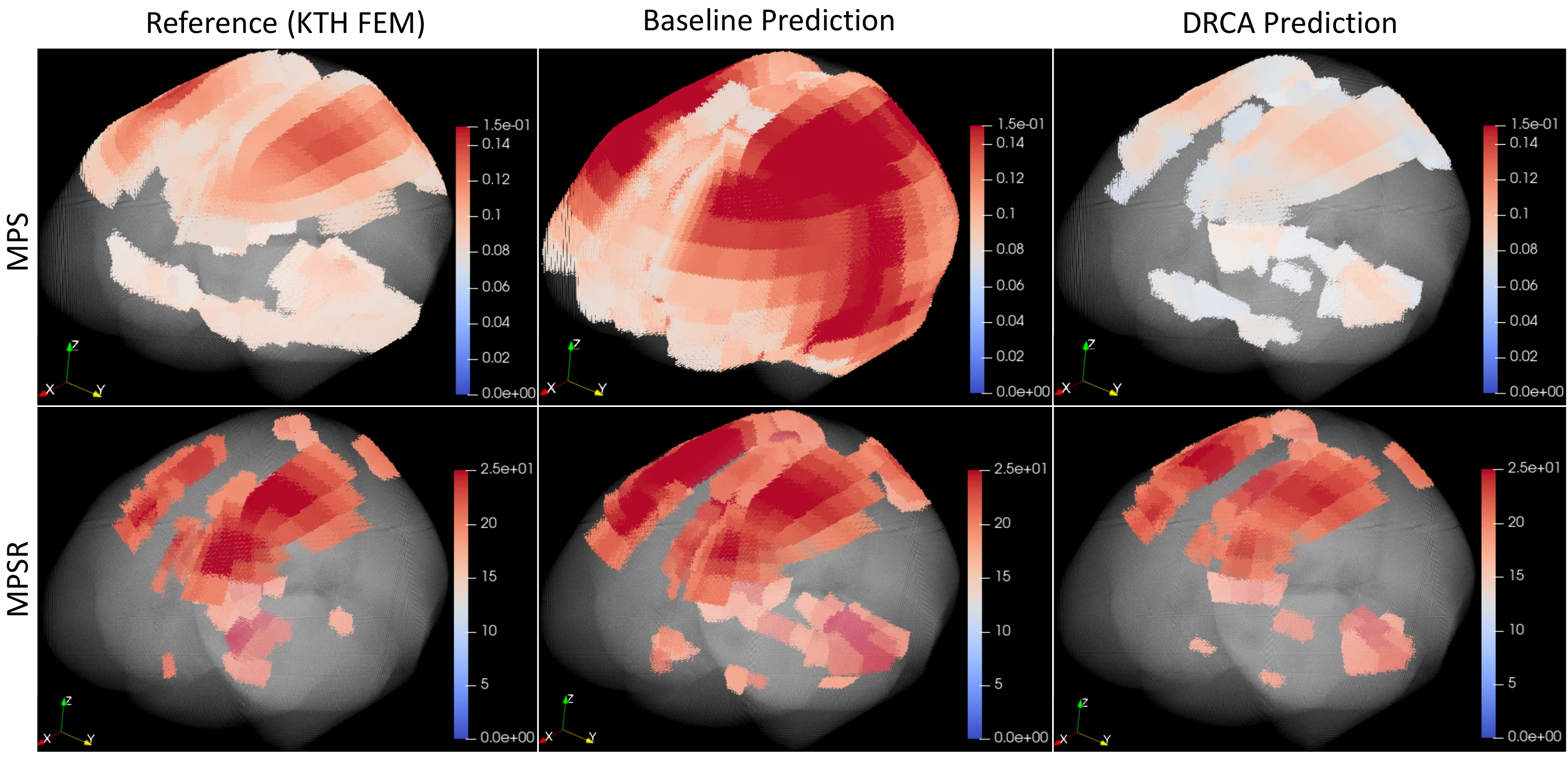}
    \caption{The 3D visualization of two example cases. Upper: MPS. Lower: MPSR. The reference values, the prediction of the baseline method and the DRCA method are shown. }
    \label{fig:4}
\end{figure*}

To visually demonstrate the prediction accuracy, we selected two example impacts and visualized the reference MPS/MPSR values, prediction values from the baseline method, and the DRCA method in Fig. \ref{fig:4}. To show the prediction of cases with relatively high brain deformation, the two example cases are at the 70th percentile after we ranked the impacts by the 95th percentile of the reference MPS/MPSR. It should be mentioned that the 95th percentile instead of the 100th percentile was used to avoid potential computational artifacts from analysis \cite{panzer2012development}. The high-strain and high-strain-rate regions predicted by the DRCA method better represented those output by the finite element modeling when compared with the predictions given by the baseline method without any domain adaptation.

% \begin{figure*}
%     \centering
%     \includegraphics[width=0.99\linewidth]{bibtex/fig/Fig3.pdf}
%     \caption{The accuracy of various unsupervised domain adaptation models on the validation set. The absolute prediction error of MPS on dataset CF1 (A) and MMA (B), the prediction error of MPSR on dataset CF1 (C) and (D) over all the brain elements and test impacts. Statistical significance of pair-wise comparison with the baseline method (paired t-test) is shown: ***: $p<0.001$}
%     \label{fig:3}
% \end{figure*}

\begin{figure*}
    \centering
    \includegraphics[width=0.99\linewidth]{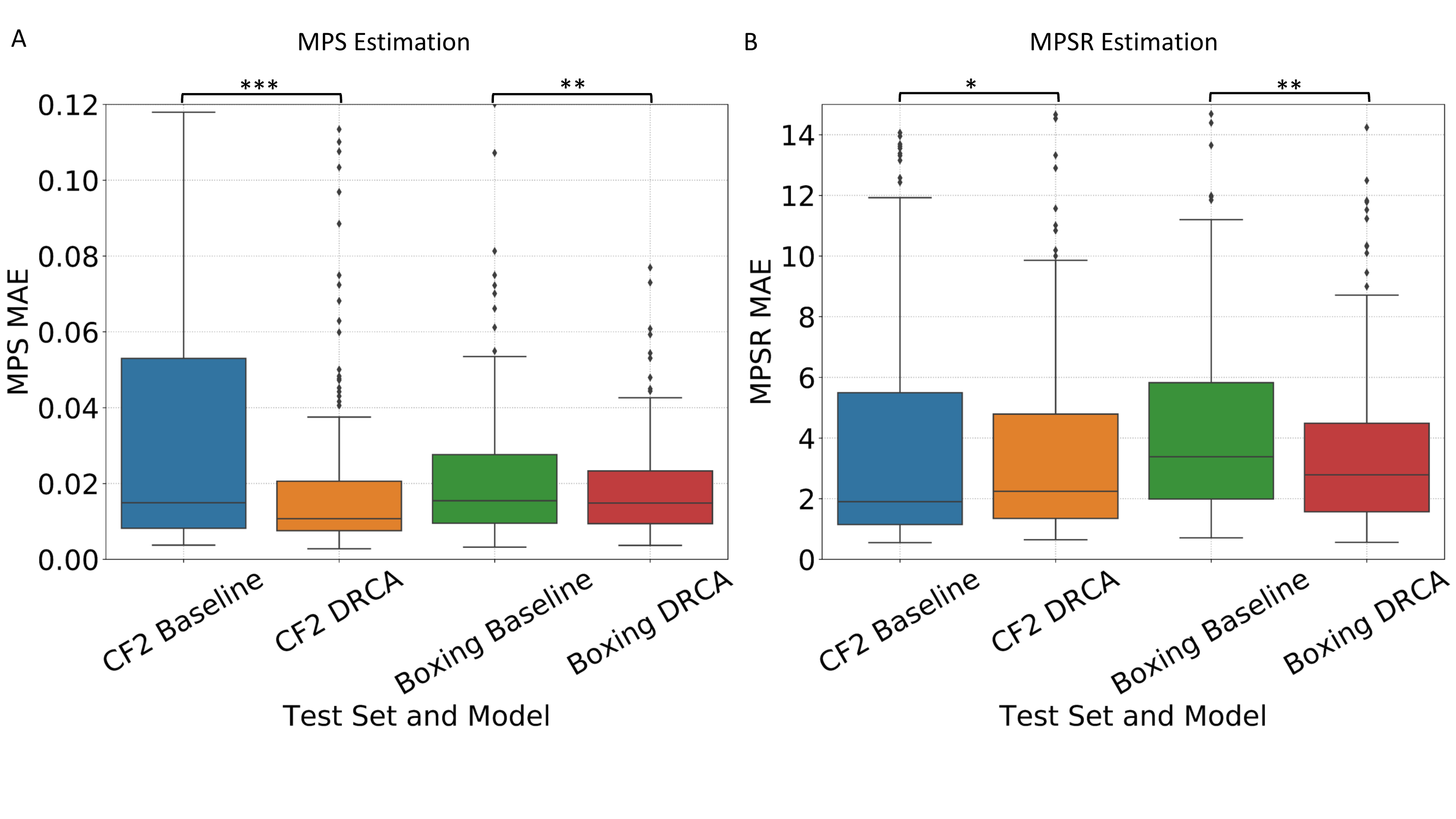}
    \caption{The accuracy of DRCA and baseline model on the two hold-out test sets. The distribution of mean absolute error over all test impacts in MPS estimation (A) and in MPSR estimation (B) on hold-out test datasets CF2 and Boxing. The MAE is computed for each test impact. Statistical significance of pair-wise comparison with the baseline method (paired t-test) is shown: *: $p<0.05$, *: $p<0.01$, ***: $p<0.001$}
    \label{fig:5}
\end{figure*}

On model performance test on the two hold-out datasets, the model accuracy was shown in Fig. \ref{fig:5}. On the dataset of 195 recently video-verified college football impacts, DRCA significantly reduced the MAE ($p \leq 0.05$) on both 1) the MPS predictio: mean MAE (over test impacts) reduced from 0.036 to 0.022 (38.9\% relative error reduction), mean RMSE (over test impacts) reduced from 0.061 to 0.043 (29.5\% relative error reduction); and 2) the MPSR prediction task: mean MAE reduced from 6.560 to 4.629 (29.4\% relative error reduction), mean RMSE reduced from 14.744 $s^{-1}$ to 8.773 $s^{-1}$ (40.5\% relative error reduction). On the dataset of 260 boxing impacts, similar results were achieved on both 1) the MPS prediction: mean MAE reduced from 0.022 to 0.018 (18.2\% relative error reduction), mean RMSE reduced from 0.037 to 0.025 (32.4\% relative error reduction); and 2) the MPSR prediction task: mean MAE reduced from 4.576 $s^{-1}$ to 3.674 $s^{-1}$ (19.7\% relative error reduction), mean RMSE reduced from 7.475 $s^{-1}$ to 5.491 $s^{-1}$ (26.5\% relative error reduction). To sum up, the DRCA, as an unsupervised domain adaptation approach is effective in generating more accurate brain strain and strain rate estimators for the measured head impacts. 

% \begin{figure*}
%     \centering
%     \includegraphics[width=0.99\linewidth]{bibtex/fig/Fig5.pdf}
%     \caption{The accuracy of DRCA and baseline model on the two hold-out test sets. The absolute value of brain-element-wise prediction error of MPS (A) and that of MPSR (B) on datasets CF2 and Boxing. The MAE of MPS/MPSR prediction across different brain regions (C). Statistical significance of pair-wise comparison with the baseline method (paired t-test) is shown: ***: $p<0.001$}
%     \label{fig:5}
% \end{figure*}

\section{Discussion}
% 1. The problem of previous MLHMs
Machine learning head models (MLHMs) have been recently developed to function as approximators of the time-consuming finite element modeling for rapid whole-brain deformation estimation for detection and real-time intervention for TBI \cite{wu2019convolutional,zhan2021rapid,ghazi2021instantaneous,wu2022real,zhan2022find}. The MLHMs developed by previous researchers can significantly reduce the time of whole-brain strain and strain rate estimation from 7-8 hours per impact by finite element modeling to within 1 ms per impact by MLHMs. Although the previously developed models have performed well on the lab-reconstructed or simulated impacts, they suffer significantly from sharp accuracy deterioration when used on on-field datasets due to the data distribution drifts potentially caused by limited data quantity of on-field datasets, different measurement devices, the measurement noises, and varying types of head impacts \cite{zhan2021relationship,zhan2021rapid}. This has also been shown in the present study: for example, without any domain adaptation, on dataset MMA, the MPS and MPSR prediction RMSE was 1.582 and 130,157.7, which were even greater than the injury thresholds (MPS: 0.3-0.4; MPSR: 120 $s^{-1}$ \cite{patton2015biomechanical,ho2007dynamic,kleiven2007predictors,hajiaghamemar2020embedded}). Therefore, the models cannot be used for brain deformation estimation for MMA impacts.

 % 2. Solution of domain adaptation in this study
To address this issue, this study develops a two-stage model configuration combining unsupervised domain adaptation with MLHMs, which manifests that with domain adaptation methods based on DRCA, the accuracy of brain strain and strain rate estimation for target datasets can be evidently improved, which is validated on four datasets of on-field college football head impacts, mixed martial arts impacts and boxing impacts. Particularly, with DRCA, the model accuracy can be significantly reduced to be even closer to the prediction performance on the simulated impacts (MPS MAE 0.015, MPSR MAE 2.818 $s^{-1}$ \cite{zhan2021rapidly}), which is much smaller than 1/10 of proposed concussion (clinically defined mTBI) thresholds \cite{kleiven2006evaluation,hajiaghamemar2020embedded}. It should be noted that this level of model accuracy cannot be achieved in previous publications of MLHMs which use the data fusion strategy to combine the simulated impacts with some on-field impacts. 

% 3. Advantage of unsupervised domain adaptation over transfer learning
Furthermore, in this study, we used the strategy of unsupervised domain adaptation. Based on the DRCA and cycle-GAN, we only used the target domain data in an unsupervised manner. Unlike transfer learning which requires users to compute the reference MPS/MPSR of the targeted impacts (on-field impacts from different types of head impacts, measured by sensors) for the fine-tuning of pre-trained models developed on the large simulated impact datasets, the domain adaptation approaches used in this study do not necessitate any labels on the target domain (on-field data). Because the function of MLHMs is to rapidly and accurately estimate brain strain and strain rate as a substitute for finite element modeling in real-world applications, the users of the MLHMs either should not have access to the finite element models to compute the ground-truth MPS and MPSR, or their application cannot rely on the finite element models to estimate whole-brain strain and strain rate due to the large computational cost. With only the head kinematics measured by wearable sensors, the users cannot leverage the transfer learning to fine-tune the pre-trained MLHMs to get better accustomed to the data distribution of a specific type of on-field impact dataset. In this case, the unsupervised domain adaptation methods investigated in this study can benefit the users by accurately estimating the whole-brain strain and strain rate caused by the on-field impacts. Therefore, research labs, clinicians and sports teams who do not have access to the resources to compute the reference MPS and MPSR can also apply MLHMs in their work. The users only need to input their impact datasets into the two-stage model and the model will perform the domain adaptation and output the strain and strain rate.

% 4. Method discussion: why domain adaptation works?
The domain adaptation methods we used in this study also warrant further discussion. The general assumption of domain adaptation methods in this study is that we assume that the domain drift is not relevant to the estimation of whole-brain strain and strain rate. When developing MLHMs, the models learn the mapping from the kinematics features to the labels (MPS/MPSR). What the models are learning involves both the brain physics (the ground-truth mapping from the domain-invariant kinematics features to the strain and strain rate) and the dataset-specific characteristics associated with each type of head impacts (simulated impacts, college football impacts, mixed martial arts impacts and boxing impacts). In order to mitigate the influence exerted by the dataset-specific characteristics, we performed the domain adaptation to minimize the influence of the domain drift but keep the domain-invariant information in the features. Ideally, the domain-invariant relationship between the kinematics features and the brain strain/strain rate is learned by the MLHMs after domain adaptation, while the dataset-specific characteristics get discarded and the domain drifts are compensated before the development of MLHMs.

% 5. Method discussion: why DRCA and cycle-GAN works?
To achieve the domain adaptation goals, we applied DRCA and cycle-GAN-based methods in this study. Firstly, as is reflected by the objective function shown in Eqn. (1), the DRCA method finds a linear transform to project the data from the original high-dimensional feature space to a lower-dimensional feature space, on which the between-domain scatter is minimized while the within-domain scatter is maintained. We propose that the reason it works well is that the minimization of between-domain scatter enables the different data distribution across domains to be mitigated. Meanwhile, the maximization of the within-domain scatter maintains the information that is going to be used to train the models to learn the relationship between the domain-invariant kinematics features to the strain and strain rate. 

Secondly, the cycle-GAN-based approaches take a different path to solve the domain adaptation issue: instead of finding a regularized lower-dimensional feature space between source domain and target domain, they directly translate the target domain to the source domain via a domain-translator based on generative adversarial networks. The domain discriminator enables the evaluation of the translated data, which drives the domain generator to be better at generating the source domain data based on the target domain data. According to the pipeline shown in Fig. \ref{fig:2}, the target domain data (on-field impacts) gets transferred to the source domain, where we can leverage the MLHMs trained on the source domain (simulated impacts). 

As a result, generally, both types of approaches work in improving the estimation of strain and strain rate on the target domain data. We deem that the comparison among these domain adaptation approaches is fair since all of them leverage the features of the simulated impacts and target impacts while the subsequent development of the MLHMs only used the labels on the simulated impacts. We hypothesize that the DRCA performs better than cycle-GAN-based approaches in that small datasets, particularly the small on-field datasets, may not enable generalizable information of each data distribution to be learned by the powerful cycle-GAN-based approach. The cycle-GAN-based approach may overfit to the large quantities of the source domain data and there may be a loss of information of the target domain data after they get transferred to the source domain. On the contrary, the DRCA is a less flexible and less complicated domain adaptor in terms of model parameters and degrees of freedom when compared with the cycle-GAN-based approaches. DRCA simply computes the between-domain scatter and within-domain scatter based on the statistical estimation of expectation and variance on the two domains. Therefore, the DRCA method is less likely to overfit to the source domain data and enables the domain-invariant features to be kept and used to train the MLHMs.

% 6. Limitations
Although this study manifests the effectiveness of using domain adaptation to improve the generalizability of MPS and MPSR estimation for on-field head impacts, there are several limitations. Firstly, in this study, the domain adaptation was done in the unsupervised manner and we regarded all the on-field impacts as unlabeled data. Therefore, we tuned the hyperparameters of the domain adaptation approaches and reported the best results on the target domain data. In the future, with the collection of more on-field data, more independent new datasets of CF impacts and MMA impacts solely used to assess the model performance are warranted to ensure the models did not overfit. Secondly, in the development of the combination domain adaptation approaches, i.e. shift-GAN and GAN+DRCA, we tuned the hyperparameters of KMM and DRCA based on the best-performing cycle-GAN, which restricted the degrees of freedom. It is also likely that these two approaches can be further optimized if the cycle-GAN is deemed as another flexible variable.

%%%%%%%%%%%%%%%%%%%%%%%%%%%%%%
\section{Conclusion}
The decreasing accuracy across diverse types of head impacts hinders the application of machine learning head models in TBI detection due to the data distribution drifts in different impact datasets. In this work, we are the first to propose an adaptive brain deformation estimator, a two-stage model configuration that integrates domain adaptation with a deep-learning brain deformation prediction network. To perform domain adaptation, we apply either a cycle-GAN-based architecture or a non-deep-learning method, DRCA, and find that DRCA achieves a substantially higher accuracy as quantified by the mean absolute error and root mean square error of the brain strain and strain rate estimation. Together, our results indicate that domain adaptation can greatly improve the accuracy of brain deformation estimation for different types of on-field head impact data and thus solve the generalizability problem of MLHMs. This study will significantly enable fast and accurate brain deformation estimation for TBI detection in potential clinical applications and for better protection of soldiers, contact-sports players and motorists who are frequently at the risks of sustaining TBI.

\section{Code and Data Availability}
The code and data associated with study can be found on \url{https://github.com/xzhan96-stf/drca-mlhm}.

\section{Declaration of conflict of interest}
The authors declare no conflict of interest.

\ifCLASSOPTIONcaptionsoff
  \newpage
\fi

\bibliographystyle{IEEEtran}
\bibliography{cite}

\end{document}